\documentclass[runningheads]{llncs}
\usepackage{graphicx}

\usepackage{tikz}
\usepackage{comment} 
\usepackage{amsmath,amssymb} 
\usepackage{color}

\usepackage{url}
\usepackage{multirow}
\usepackage{subfigure}
\usepackage{threeparttable}

\begin{document}
\pagestyle{headings}
\mainmatter
\def\ECCVSubNumber{3}  

\title{A Flow Base Bi-path Network for Cross-scene Video Crowd Understanding in Aerial View} 

\titlerunning{A Flow Base Bi-path Network for Crowd Understanding}
%
\author{Zhiyuan Zhao\protect\footnote{Equal Contribution\protect\label{equcon}} \and
Tao Han\textsuperscript{\ref{equcon}}  \and
Junyu Gao \and
Qi Wang\protect\footnote{Corresponding Author}  \and 
Xuelong Li}
\authorrunning{Z. Zhao et al.}
%
\institute{School of Computer Science and Center for OPTical IMagery Analysis and Learning (OPTIMAL), Northwestern Polytechnical University, Xi’an 710072, Shaanxi, P.R. China.
\email{tuzixini@163.com,hantao10200@mail.nwpu.edu.cn,\\\{gjy3035,crabwq\}@gmail.com,li@nwpu.edu.cn}
}
\maketitle

\begin{abstract}
  Drones shooting can be applied in dynamic traffic monitoring, object detecting and tracking, and other vision tasks. The variability of the shooting location adds some intractable challenges to these missions, such as varying scale, unstable exposure, and scene migration. In this paper, we strive to tackle the above challenges and automatically understand the crowd from the visual data collected from drones. First, to alleviate the background noise generated in cross-scene testing, a double-stream crowd counting model is proposed, which extracts optical flow and frame difference information as an additional branch. Besides, to improve the model's generalization ability at different scales and time, we randomly combine a variety of data transformation methods to simulate some unseen environments. To tackle the crowd density estimation problem under extreme dark environments, we introduce synthetic data generated by game Grand Theft Auto V(GTAV). Experiment results show the effectiveness of the virtual data. Our method wins the challenge with a mean absolute error (MAE) of 12.70\footnote{Finally, reach the MAE of 12.36, ranked the second.\protect\label{final}}. Moreover, a comprehensive ablation study is conducted to explore each component's contribution.
\end{abstract}

\section{Introduction}

  Video analysis \cite{borges2013video,zhang2018physics} has become an increasingly important part of computer vision, which involves a wide range of fields, such as object detection and tracking \cite{lin2014microsoft,bertinetto2016fully}, crowd segmentation \cite{kang2014fully}, density estimation and localization \cite{idrees2018composition,wen2019drone}, and group detection \cite{wang2018detecting}. Among them, a new challenge that understands the crowd from a drone perspective has recently attracted lots of attention. Compared with the density estimation in the ground scenes, unmanned aerial vehicles (UAVs) have a broader surveillance scope. It can play a more crucial role in public safety and urban management.

  The scale of the datasets limited the development of crowd counting tasks over the past few years. However, it has been booming driven by the release of a series of large-scale datasets recently, such as UCF-QNRF\cite{idrees2018composition}, JHU-Crowd \cite{sindagi2020jhu}, NWPU-Crowd \cite{wang2020nwpu}. Unlike the counting task from the ground perspective, there are extra challenges in the UAV scenarios. In this paper, we work on density and crowd estimation for the DroneCrowd dataset \cite{wen2019drone}. As illustrated in Fig. \ref{fig:intro}, we found that the difficulties of UAV's task of counting people lie in the following three aspects:

  \begin{figure}[t]
    \centering
    \includegraphics[width=0.98\textwidth]{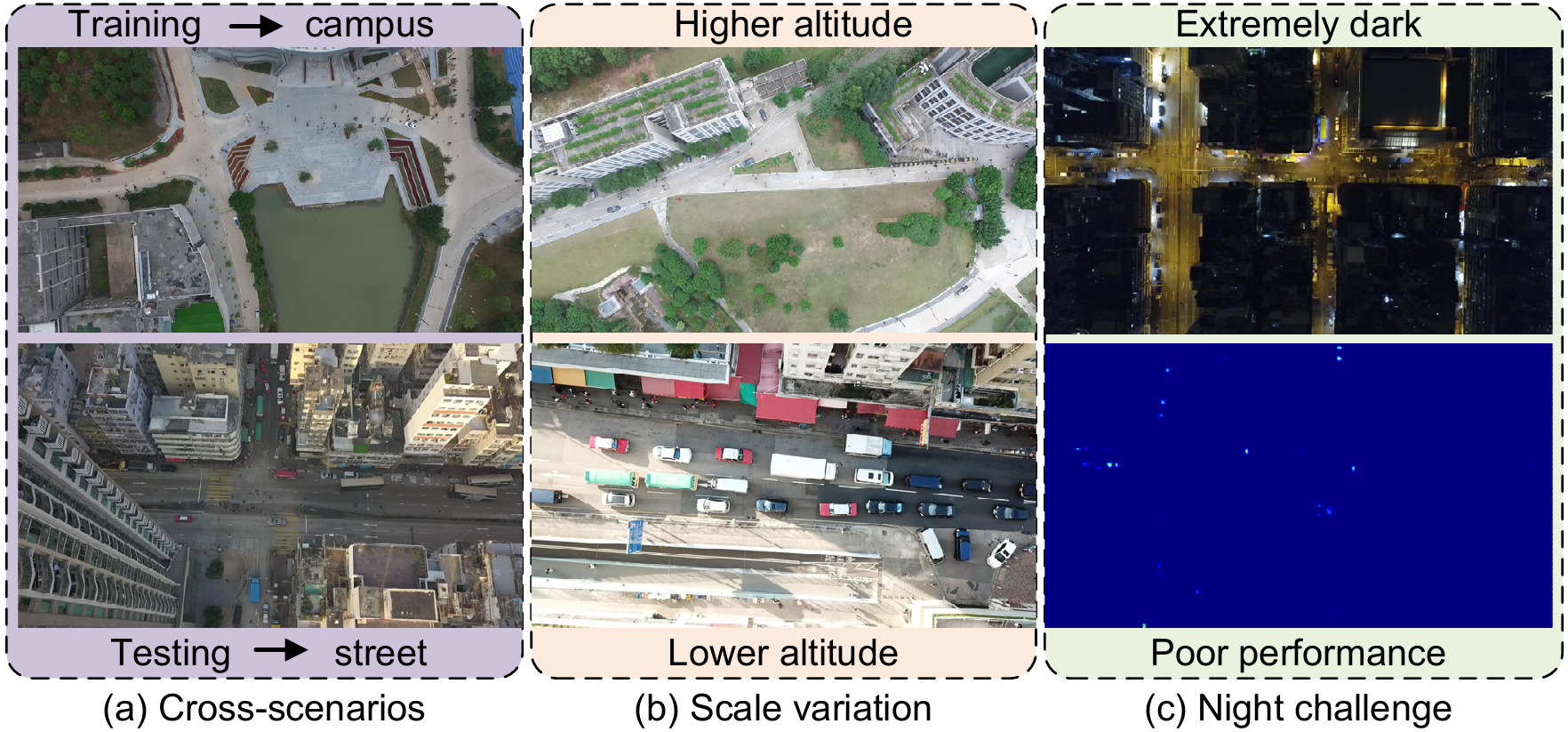}
    \caption{Three challenges in the DroneCrowd Dataset.}
    \label{fig:intro}
  \end{figure}

  \begin{itemize}
    \item \emph{Cross-scenarios}. The training set and validation set of the DroneCrowd dataset are divided according to the scenario, which means that it is a cross-scene crowd counting task. The test set scenarios are almost unknown to the model, which increases the difficulty of the task.
    \item \emph{Scale variation}. The flying height of drones is constantly changing during data collection, which significantly affects the scales of objects. The diversity of scale increases the difficulty of model fitting.
    \item  \emph{Night challenge}. There are some extremely dark scenes in the test set, but there are no scenes with similar illumination conditions in the training set. These night scenarios have a significant impact on the test results.
  \end{itemize}

  The biggest problem of cross-scene attributes is that it has serious background noise in testing unknown scenes, which is impossible to be tackled directly by existing counting models. The effective measure is to reduce background noise by enhancing the model's ability to distinguish between background and moving objects. Note that the DroneCrowd dataset is composed of video sequences. We can extract the optical flow information of the video to achieve this purpose. Therefore, in our counting framework, an additional branch is designed to input optical flow information to enhance the feature extraction of moving objects. Besides, to improve the generalization ability of the model for the scale variables, multi-scale images are transferred to simulate the scenes at different heights. Finally, to alleviate the large test deviation caused by low illumination scenes, a variety of image brightness adjustment algorithms and synthetic data are used to simulate different illumination conditions in the training sequences. All three ideas contribute to the competitive performance, which we will explore in detail in section \ref{sec:exp}.

  As a summary, we propose a bi-path framework based on RGB image and optical flow information to improve the model's ability to weaken the background noise for unknown scenes and exploit multi-scale transformation and multi-luminance adjustment algorithms to make data augmentation for reducing the adverse effects of scale and illumination. 
  Besides, we generate synthetic data under dark scenes, which expands training data using low-cost methods and improves the performance of the module in extreme scenarios.

\section{Related Work}

  \quad\quad\textbf{Scale-aware Crowd Counting.}\quad
  With the development of crowd counting, it is found that scale change has the greatest negative impact among the influencing factors (occlusion, scale, and viewpoint, etc.). Many crowd counting algorithms focus on scale variability in recent years. Zhang \emph{et al.} \cite{zhang2016single} propose a three-columns network with different kernels for scale perception in 2016. Onoro-Rubio \emph{et al.} \cite{onoro2016towards} introduce a Hydra CNN with three-columns, where each column is fed by a patch from the same image with a different scale. Wu \emph{et al.} \cite{wu2019adaptive} develop a powerful multi-column scale-aware CNN with an adaptation module to fuse the sparse and congested columns. In the same year, Adaptively Fusing Predictions (AFP) \cite{kang2018crowd} generates a density map by fusing the attention map and intermediary density map in each column. ic-CNN \cite{ranjan2018iterative} generates a high-resolution density map via passing the feature and predict map from the low-resolution CNN to the high-resolution CNN. SDA-MCNN \cite{yang2019counting}, a scale-distribution-aware neural network that resolves scale change by processing a crowd image with multiple Convolutional Neural Network columns and minimizing the per-scale loss. Hossain \emph{et al.} \cite{hossain2019crowd} employs a scale-aware attention network, where each column is weighted with the output of a global scale attention network and local scale attention network. ACM-CNN \cite{zou2019attend}, Adaptive Capacity Multi-scale convolutional neural networks assign different capacities to different portions of the input based on three modules: a coarse network, a fine network, and a smooth network. Except for multi-column scale-aware architecture, some works focus on the single-column scale-aware CNN, such as SANet \cite{cao2018scale}, SaCNN \cite{zhang2018crowd}. To combine the multi-column and single-column scale-aware CNN, CSRNet\cite{li2018csrnet}, CAN \cite{liu2019context} and FPNCC \cite{cenggoro2019feature} develop a model containing multiple paths only in several part of the networks. In 2020, SCAN \cite{yuan2020crowd}, SRN+PS \cite{dong2020scale}, SRF-Net \cite{chen2020scale}, and ASNet \cite{jiang2020attention} also explore the scale-aware to improve the counting performance.
  
  \textbf{Cross-scene/domain Crowd Counting.}\quad
  Due to the laborious data annotation required of crowd datasets, cross-scene, and cross-domain crowd counting attract researchers' attention in recent years. In this task, the model is trained on a labeled dataset and then adapted to an unseen scene. Authors in \cite{Zhang2016data} establish the earliest cross-scene dataset, which includes 103 scenes for training and the remaining five scenes for testing, a total of 3980 images. In \cite{liu2016cross}, they propose a Fully Convolutional Neural Network(FCN) and a weighted adaptive human Gaussian model for person detection and then apply it to the new scene with few labeled data. DA-ELM \cite{yang2018cross}, a counting model based on domain adaptation-extreme learning machine, counts the people in a new scene with only a half of the training samples compared with counting without domain adaptation. In \cite{hossainone}, they propose a one-shot learning approach for learning how to adapt to a target scene using one labeled example. In \cite{reddy2020few}, they apply the MAML \cite{finn2017model} to learn scene adaptive crowd counting with few-shot learning. Inspired by the synthetic data can automatically label as the source domain, Wang \emph{et al.} \cite{wang2019learning} launch a large-scale synthetic dataset to pre-train a model and adapt it to real-world datasets by a fine-tune operation. Except fine-tuning, they also complete counting without any real-world labeled information by using the Cycle GAN \cite{zhu2017unpaired} and SE Cycle GAN \cite{wang2019learning} to generate realistic images. Recently, several efforts have been made to follow them. DACC \cite{gao2019domain}, a method for domain adaptation based on image translation and Gaussian-prior reconstruction, achieves a better performance on several mainstream datasets. At the same time, some works\cite{gao2019feature,han2020focus,nie2019towards,wang2020discriminative} extract domain invariant features based on adversarial learning. In \cite{wang2020neuron}, the authors propose a Neuron Linear Transformation (NLT) method that models domain differences and uses few labeled target data to train the domain shift parameters. Experimental results show that cross-domain crowd counting can alleviate the problem of data annotation to some extent.

\section{Proposed Method}

A perfect crowd counter should have stable and excellent performance in practical applications. 
However, as mentioned above, the actual application scenarios are complex and changeable, and the training data is limited.
Therefore, in our proposed method, we promote the model's performance from the following two aspects:
\begin{itemize}
  \item Improve the feature extraction and representation ability of the crowd count network through better network structure design. Therefore, it boosts the cross-scene stability and accuracy of the model.
  \item Try to mine the information contained in the limited training data. This includes using different kinds of data transform methods to manually create unseen verity scenario from existing data, and make use of the joint information between adjacent frames while reading continuous sequential data.
\end{itemize}
The remaining paragraphs follow these two aspects to introduce the work done by the proposed method.

\subsection{Network Design}
\begin{figure}[t]
    \centering
    \includegraphics[width=0.98\textwidth]{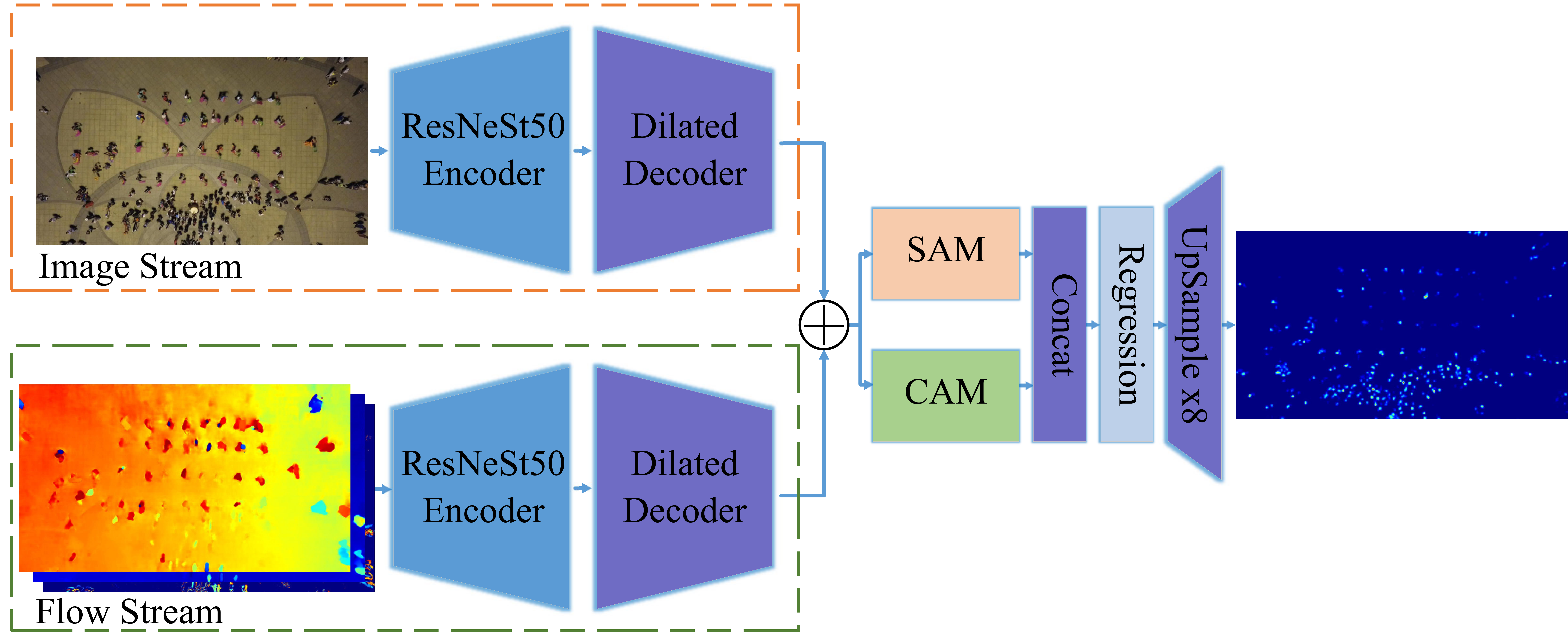}
    \caption{Network structure of the proposed bi-path crowd counter.}
    \label{fig:fra}
  \end{figure}

As shown in Figure~\ref{fig:fra}, we design a bi-path flow-based crowd counting network to make use of inter-frame information together with the input RGB image.
Each pathway consists of one powerful feature extractor and several convolutional layers as the decoder.
The encoder is the the first three layers of the newly proposed ResNeSt\cite{zhang2020resnest} with weights pre-trained on the ImageNet\cite{deng2009imagenet} dataset.
The crowd under the aerial view is small, and the building area is large. 
To better analyze the relationship between humans and background, we follow the idea of CSRNet\cite{li2018csrnet}, the decoder contains six dilated convolutions, which reduces the number of channels while parsing the semantic information of the input features.
Atrous convolution layers enlarge the convolution kernel's reception field, which enriches spatial contextual information inside the features.
The decoder outputs feature vectors of 1/8 size of the original input with 64 channels.
The image branch and the flow branch are combined immediately after the decoders.
Besides the dilated convolutions, we introduce spatial-wise attention module (SAM) and channel-wise attention module(CAM) from \cite{gao2019scar} to enhance large-range dependencies on the spatial and channel dimensions.

  The detailed network structure is given in Table~\ref{tab:str}. 
  ``ResNeSt50\_Conv1\_c64'' means this layer has the same structure as the Conv1 layer of ResNeSt50, and its output is a 64 channel tensor.
  ``k(3,3)-c512-s1-d2-R'' represents convolutional layer with kernel size $3\times3$,output channel 512, stride 1, dilation rate 2, and followed by a ReLU activation layer.

  \begin{table}[t]
    \centering
    \caption{Network structure of the proposed method.}
      \begin{tabular}{c|c}
        \hline
        \multicolumn{2}{c}{\textbf{Encoder}}         \\\hline
        \multicolumn{2}{c}{ResNeSt50\_Conv1\_c64}    \\
        \multicolumn{2}{c}{ResNeSt50\_Layer1\_c256}  \\
        \multicolumn{2}{c}{ResNeSt50\_Layer2\_c512}  \\
        \multicolumn{2}{c}{\ ResNeSt50\_Layer3\_c1024\ } \\\hline
        \multicolumn{2}{c}{\textbf{Decoder}}         \\\hline
        \multicolumn{2}{c}{k(3,3)-c512-s1-d2-R}      \\
        \multicolumn{2}{c}{k(3,3)-c512-s1-d2-R}      \\
        \multicolumn{2}{c}{k(3,3)-c512-s1-d2-R}      \\
        \multicolumn{2}{c}{k(3,3)-c256-s1-d2-R}      \\
        \multicolumn{2}{c}{k(3,3)-c128-s1-d2-R}      \\
        \multicolumn{2}{c}{k(3,3)-c64-s1-d2-R}       \\\hline
        \multicolumn{2}{c}{\textbf{Attention}}       \\\hline
        \ \ \ \ \ SAM \ \ \ \ \ & CAM \\\hline
        \multicolumn{2}{c}{concat}                   \\\hline
        \multicolumn{2}{c}{\textbf{Regression}}      \\\hline
        \multicolumn{2}{c}{k(3,3)-c512-s1-d2-R}      \\
        \multicolumn{2}{c}{UpSample $\times 8$}     \\\hline
      \end{tabular}
    \label{tab:str}
  \end{table}

  The image stream accepts the three-channel RGB color image as input, while the flow stream also takes a three-channel input with the same size as the original image.

  Generally speaking, the optical flow is a two-channel vector, and each channel stores the optical flow information of the image along the horizontal axis and vertical axis separately. 
  We name this form of optical flow $f_x$ and $f_y$.
  It can also be transformed into polar coordinates.
  Under the polar system, two transformed channels represent the polar radius and polar angle, respectively.
  Researchers usually take these two channels as the first two channels in the HSV color space, and then convert them back to the RGB space for visualization, so we name them $f_h$ and $f_s$ here.
  The flow stream takes a three-channel vector as input, here we fill the third dimension of the input data with the frame difference vector of the two adjacent frames. Which means take frame t subtracts frame t+1 and get $f_{sub}$.
  Different forms of optical flow contain different information, leading to distant results to the network's training. We have tried several combinations; the specific experimental results and analysis are given in Section~\ref{sec:abl}.

  \begin{figure}[tbp]
    \centering
    \begin{minipage}[t]{0.98\textwidth}

      \begin{minipage}[t]{0.328\textwidth}
        \centering
        \includegraphics[width=1\textwidth]{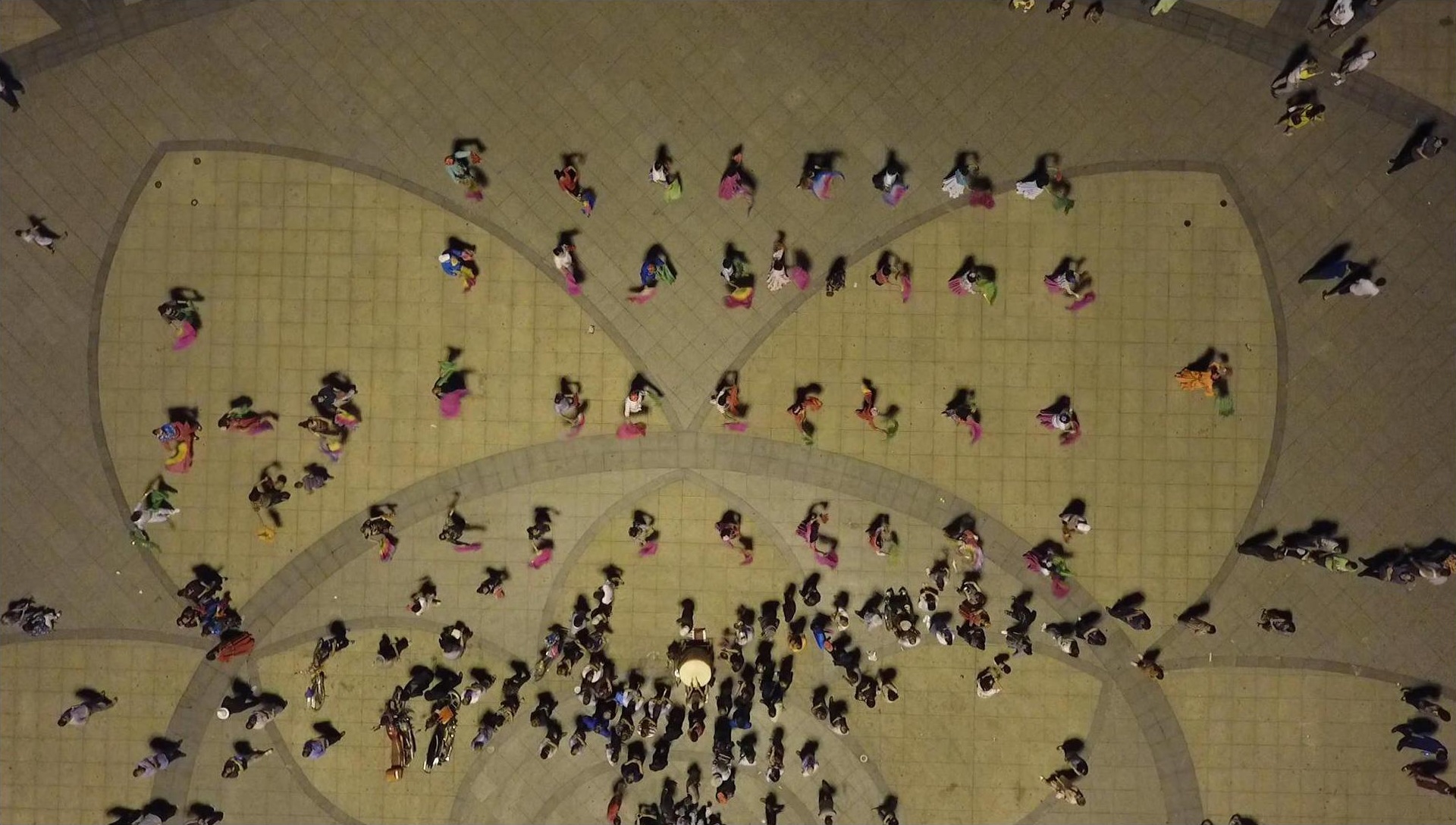}
        Original Image
      \end{minipage}
      \begin{minipage}[t]{0.328\textwidth}
        \centering
        \includegraphics[width=1\textwidth]{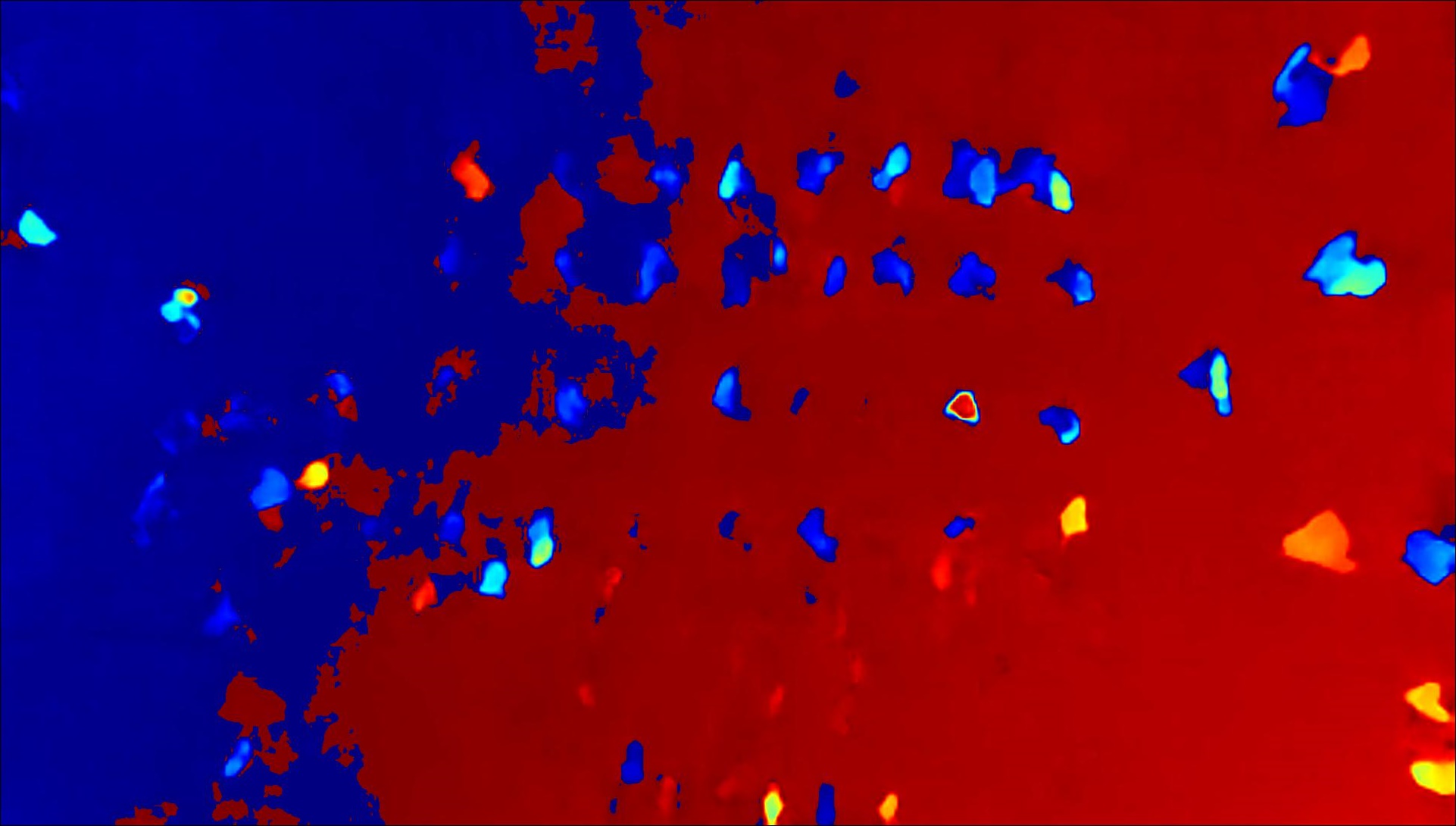}
        $f_x$
      \end{minipage}
      \begin{minipage}[t]{0.328\textwidth}
        \centering
        \includegraphics[width=1\textwidth]{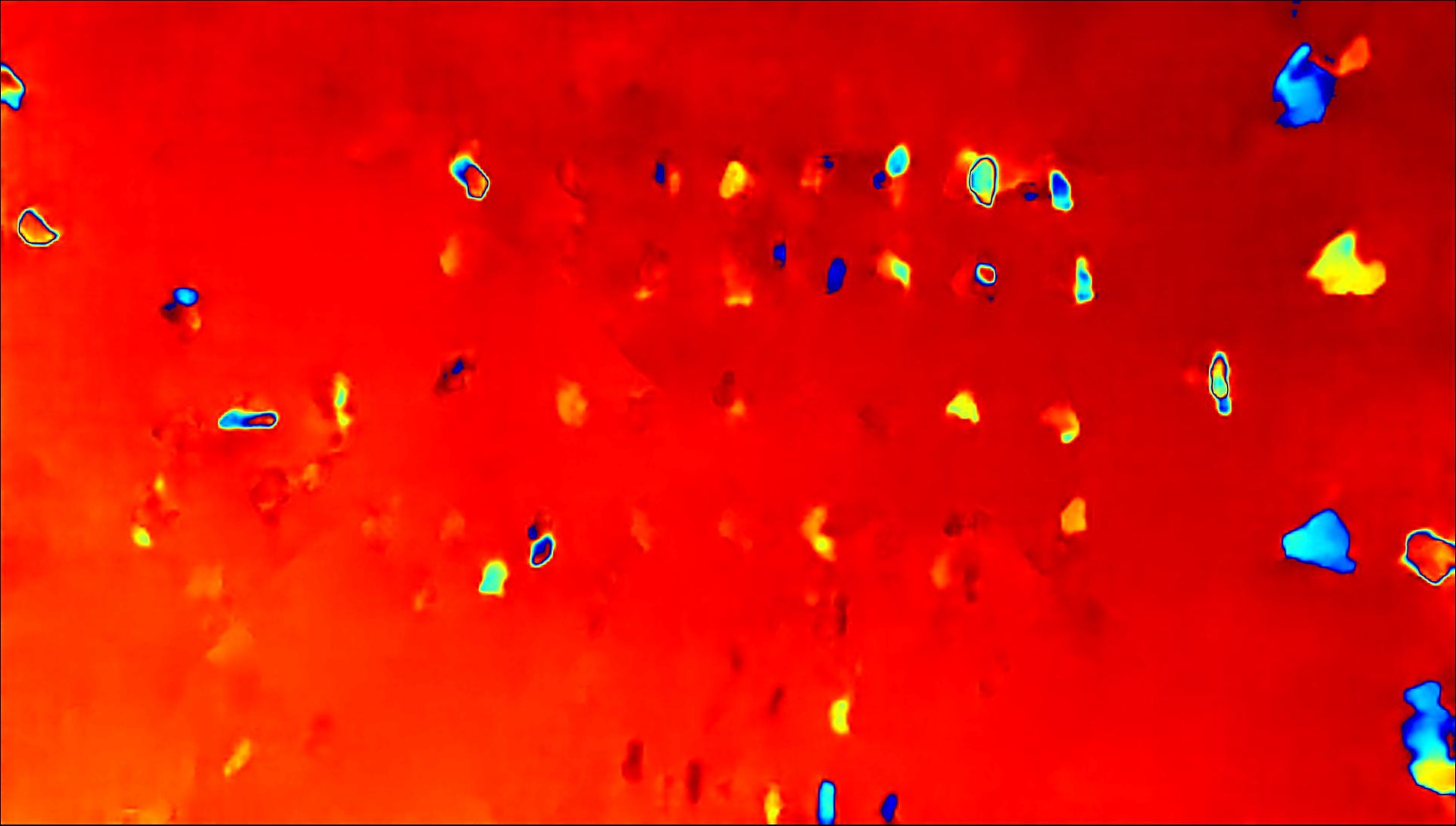}
        $f_y$
      \end{minipage}

      \begin{minipage}[t]{0.328\textwidth}
        \centering
        \includegraphics[width=1\textwidth]{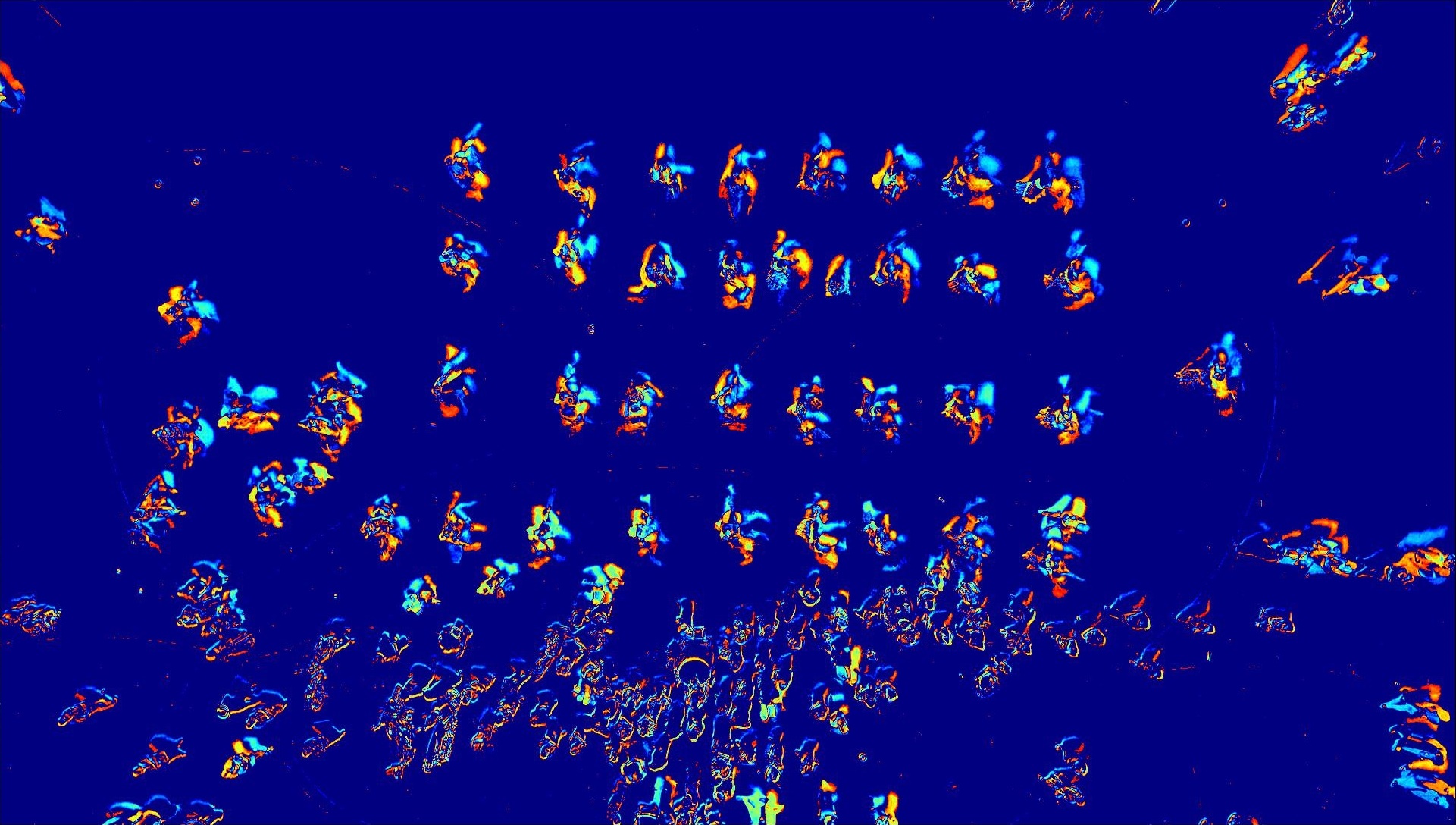}
        $f_{sub}$
      \end{minipage}
      \begin{minipage}[t]{0.328\textwidth}
        \centering
        \includegraphics[width=1\textwidth]{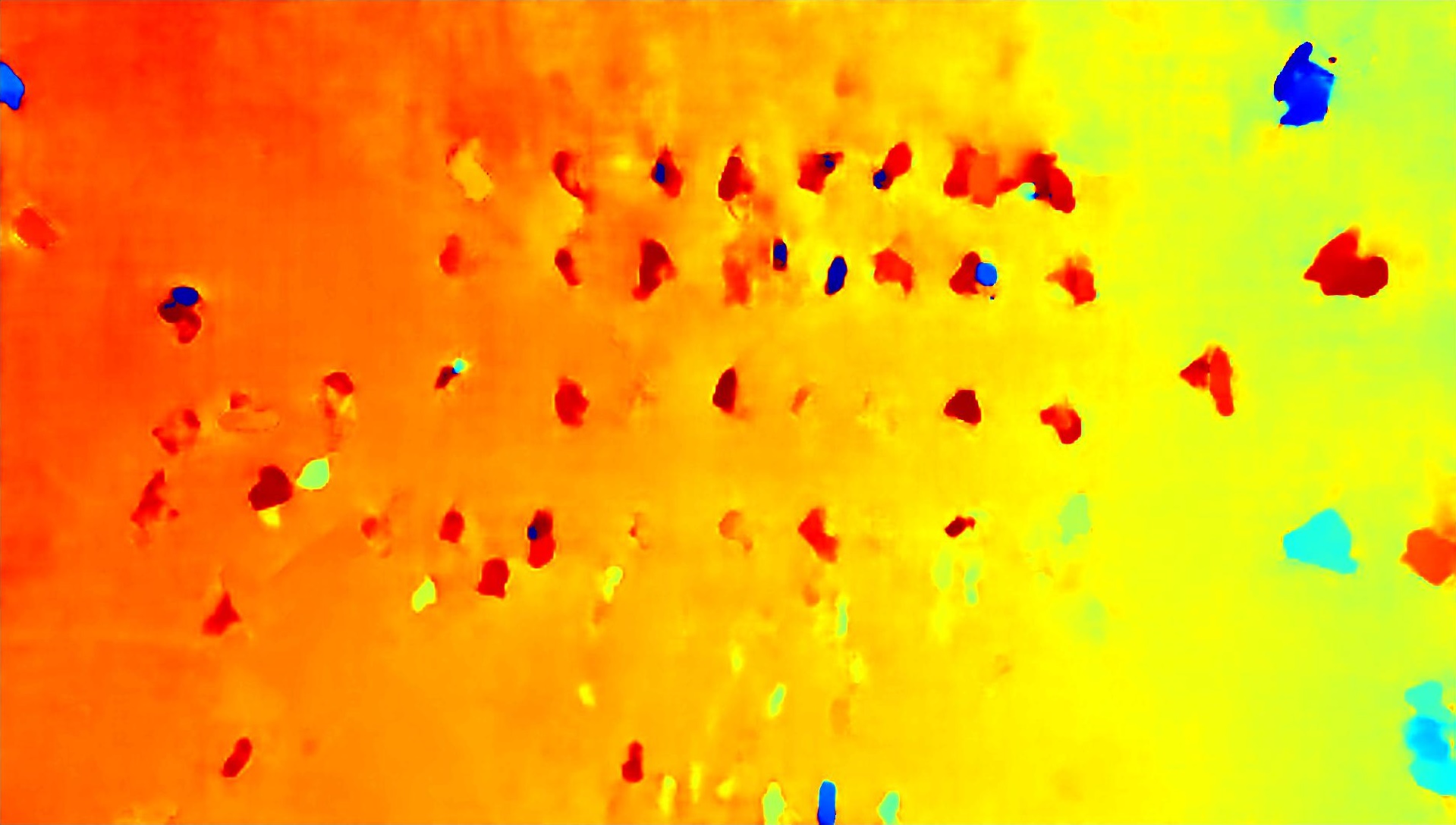}
        $f_h$
      \end{minipage}
      \begin{minipage}[t]{0.328\textwidth}
        \centering
        \includegraphics[width=1\textwidth]{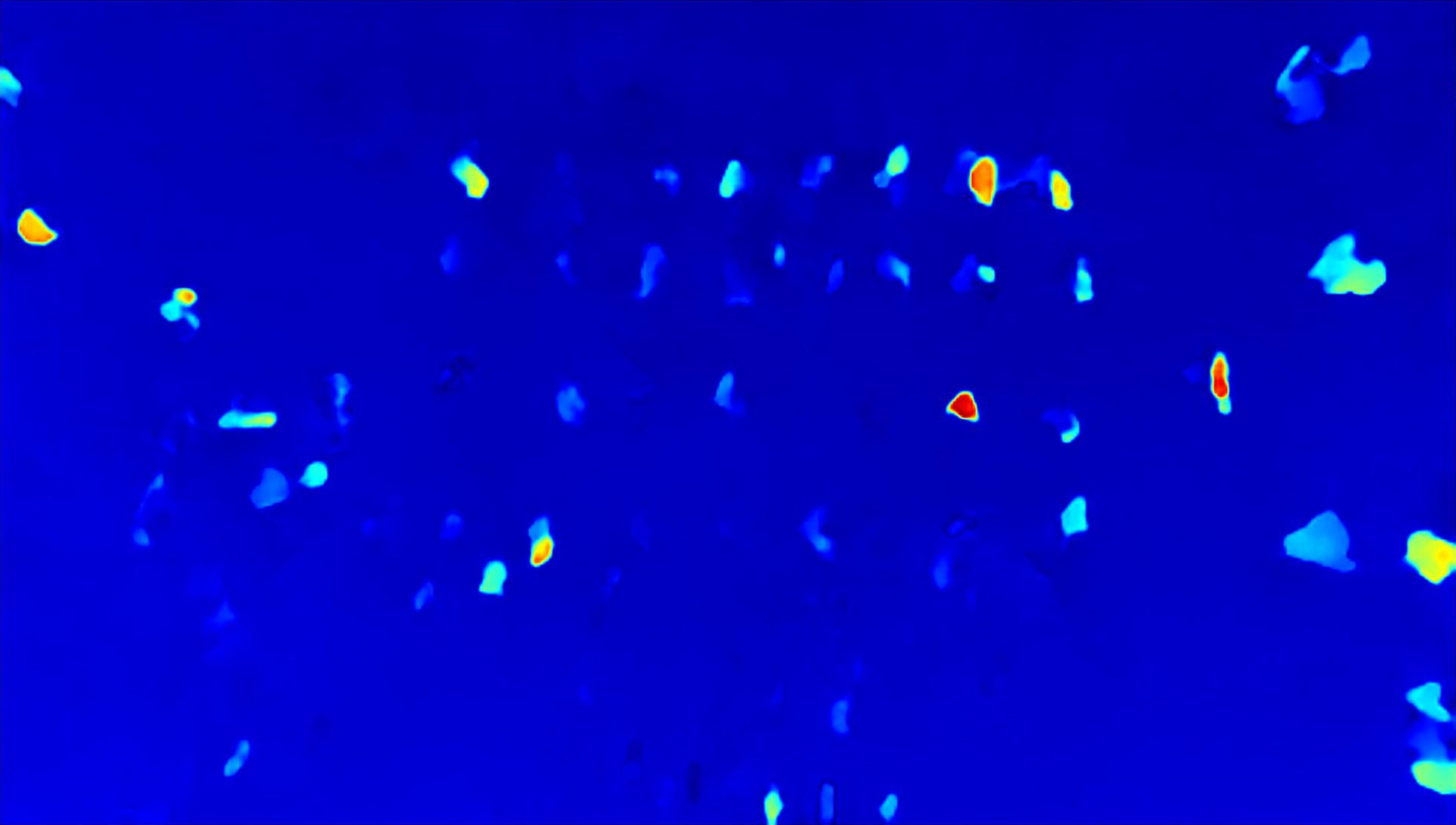}
        $f_s$
      \end{minipage}

    \end{minipage}

    \caption{Visualization of various optical flow forms.}
    \label{fig:flow}
\end{figure}

Figure~\ref{fig:flow} visualizes different forms of optical flow data. 
During the visualization and analysis process, we find that there may be a slight jitter in the process of data acquisition, which introduces a large area of noise to the optical flow information. 
Therefore, we add a simple threshold filtering in the actual use, which effectively suppresses part of the noise.

\subsection{Data Argumentation}
Apart from the network structure design, it is also significant to effectively use existing data.

\begin{figure}[tbp]
    \centering
    \begin{minipage}[t]{0.98\textwidth}

      \begin{minipage}[t]{0.245\textwidth}
        \centering
        \includegraphics[width=1\textwidth]{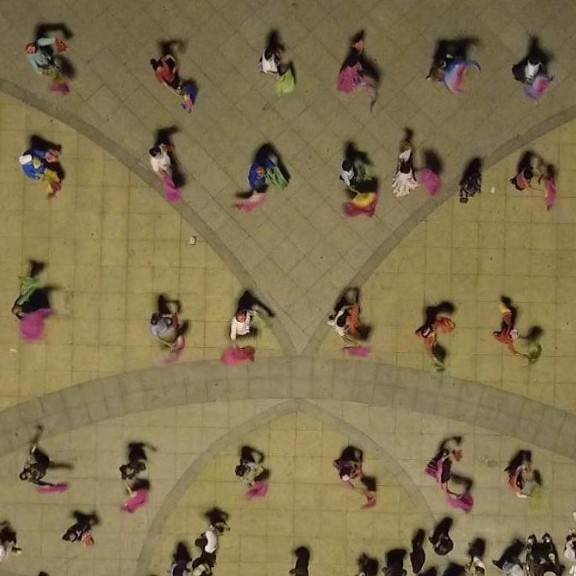}
        $\mathcal{T}_{\mathcal{C}}(\mathcal{I})$
      \end{minipage}
      \begin{minipage}[t]{0.245\textwidth}
        \centering
        \includegraphics[width=1\textwidth]{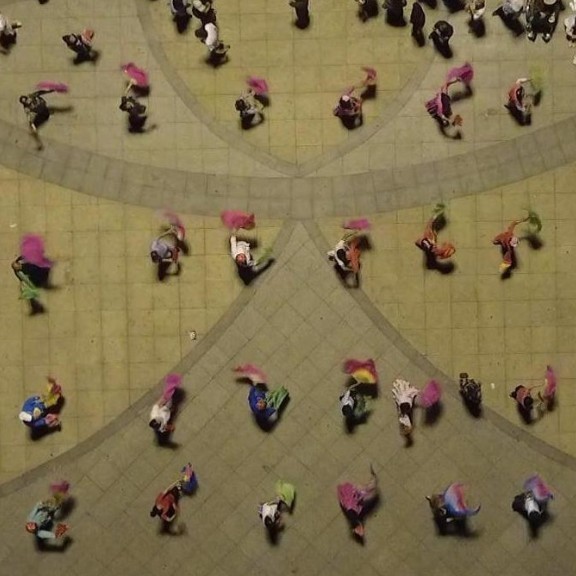}
        $\mathcal{T}_{\mathcal{CF}}(\mathcal{I})$
      \end{minipage}
      \begin{minipage}[t]{0.245\textwidth}
        \centering
        \includegraphics[width=1\textwidth]{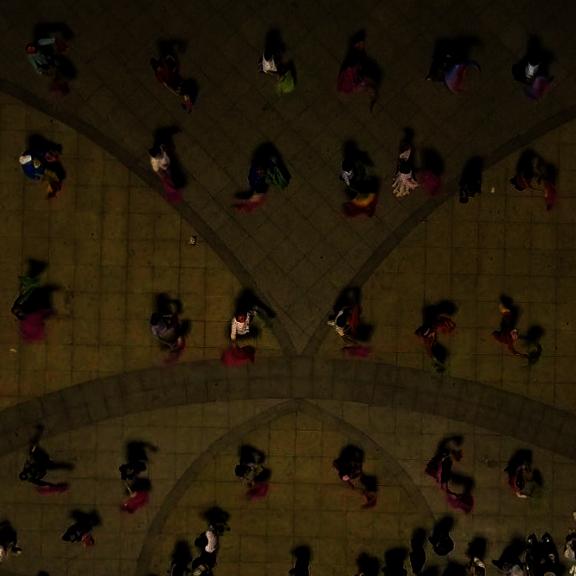}
        $\mathcal{T}_{\mathcal{CG}}(\mathcal{I})$
      \end{minipage}
      \begin{minipage}[t]{0.245\textwidth}
        \centering
        \includegraphics[width=1\textwidth]{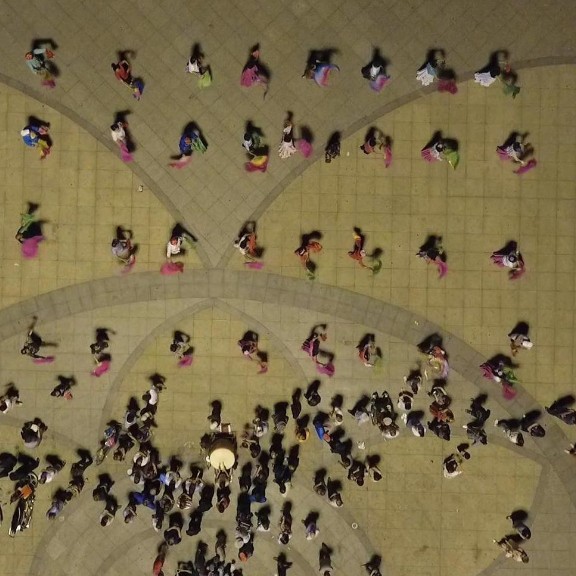}
        $\mathcal{T}_{\mathcal{CS}}(\mathcal{I})$
      \end{minipage}

    \end{minipage}
    \caption{Visualization of various data argumentation operations. The original image is the same as the one in Figure~\ref{fig:flow}. Two subscripts write together indicates the usage of two kinds of transformation together.}
    \label{fig:TF}
  \end{figure}

The training data only contains limited scenes. 
By observing, analyzing, and summarizing the status of the existing data, we speculate the scenarios that may be encountered during the actual use. 
We manually augment the existing data of the scene according to the actual situation. 
During training, numerous transform methods were used to generate new data, improving the model's generalization ability.
Instead of original data group $(\mathcal{I},F,D)$ (Image, Flow, DotMap), new training data group $(\mathcal{I}_{\mathcal{N}},F_{\mathcal{N}},D_{\mathcal{N}})$ is generated by random transform method $\mathcal{T}(\cdot)$:
  \begin{equation}
    (\mathcal{I}_{\mathcal{N}},F_{\mathcal{N}},D_{\mathcal{N}}) = \mathcal{T}_{\{\mathcal{C,F,G,S}\}}(\mathcal{I},F,D),
  \end{equation}
  where subscript $\{\mathcal{C,F,G,S}\}$ represents crop, flip, gamma correction, scale change respectively.
  The following part of this subsection introduces all data transformation methods we use according to the characteristics of the data.
  Figure~\ref{fig:TF} gives examples of several transformations.

  \emph{Scene Change: }
  Depending on the drone's shooting location, different scenes may vary considerably, whether it is the street or the park. Their style and direction will change. 
  To cope with the scenario's variability, we make random transforms to each group of input data during the training process, which includes random crop into fixed size ($576\times 576$ here), random flip (up-down and left-right).
  Of course, those transformations perform the same processing within each group of data $(\mathcal{I},F,D)$ used for training. 
  We name those transformations $\mathcal{T}_{\{\mathcal{C,F}\}}(\cdot)$.

  \emph{Extreme Light Variation: }
  All the training data we have are taken under good light conditions, but the data used in the test and the real world may not always have such good light conditions. There will be overexposure or underexposure. So before we send the data into the model, we carry out random gamma correction on the RGB image with a certain probability (here is 0.5). The range of Gamma values is from 0.4 to 2 to simulate excessive brightness and insufficient brightness.
  This transformation only applies on the RGB images, and we call it $\mathcal{T}_{\mathcal{G}}(\cdot)$.

  \emph{Scale Diversity: }
  When the UAV collects data in different states, it may fly under a low or high altitude, seriously affecting the size of the crowd in the picture and ultimately affecting the model's analysis results. 
  To further enhance the model's generalization ability, we force it to deal with images collected at various heights. 
  We change the scale of the images before cropping, which means randomly enlarge or shrink the image, and then crop the new one for training.
  This transformation works on all three input data together. We name it $\mathcal{T}_{\mathcal{S}}(\cdot)$.

  \subsection{Synthetic Data}
\begin{figure}[tbp]
      \centering
      \begin{minipage}[t]{0.98\textwidth}
  
        \begin{minipage}[t]{0.328\textwidth}
          \centering
          \includegraphics[width=1\textwidth]{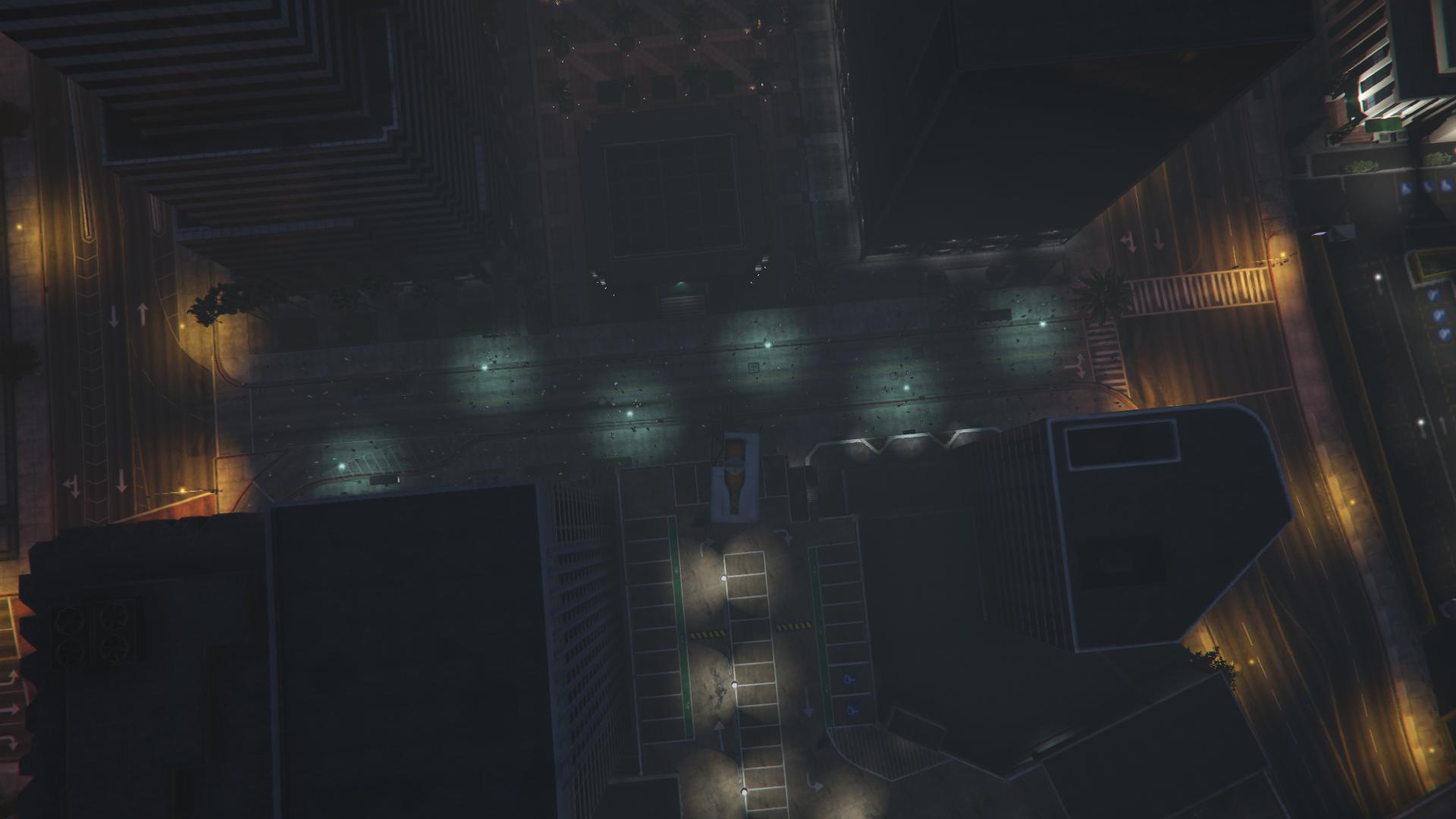}
        \end{minipage}
        \begin{minipage}[t]{0.328\textwidth}
          \centering
          \includegraphics[width=1\textwidth]{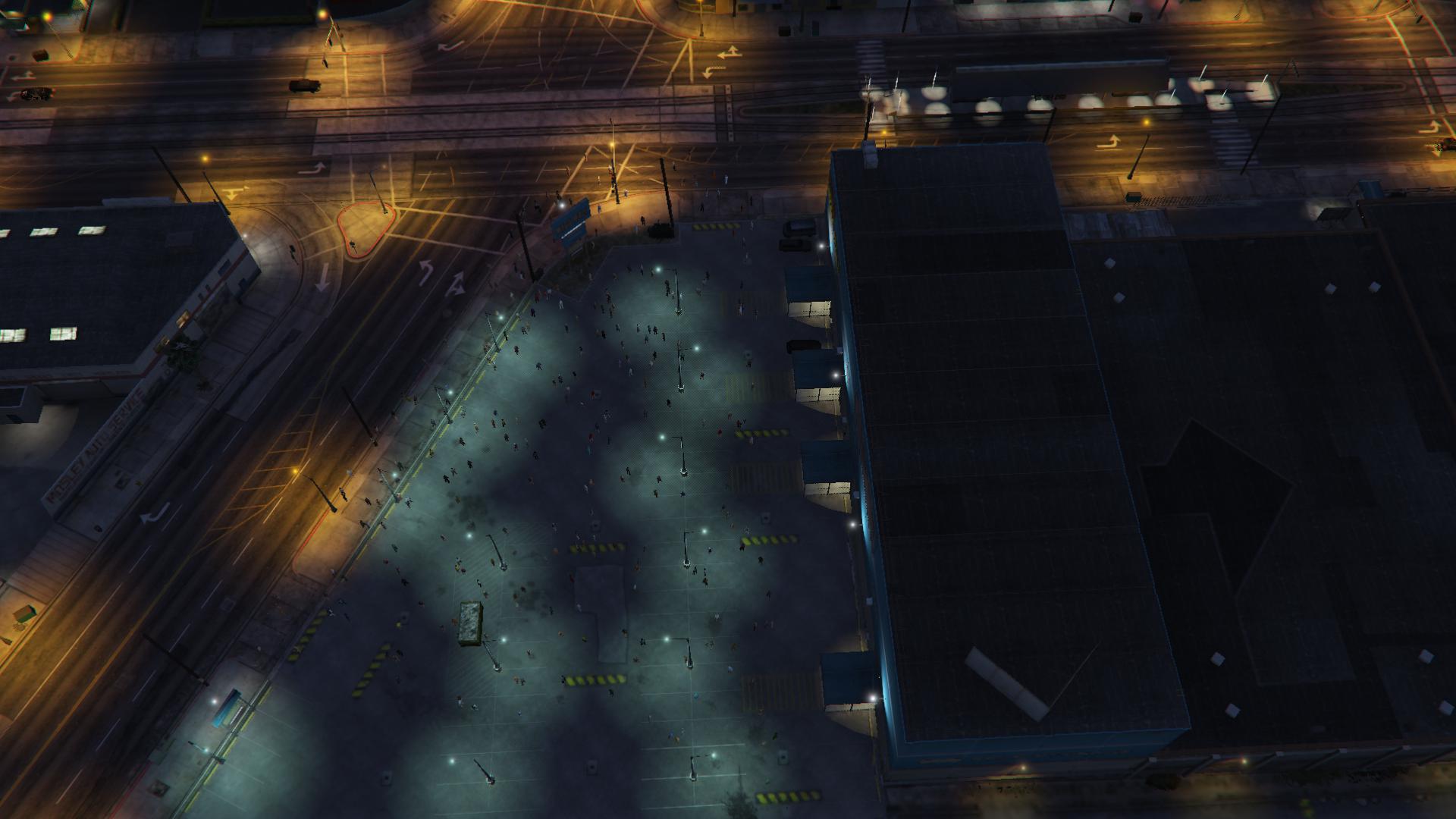}
        \end{minipage}
        \begin{minipage}[t]{0.328\textwidth}
          \centering
          \includegraphics[width=1\textwidth]{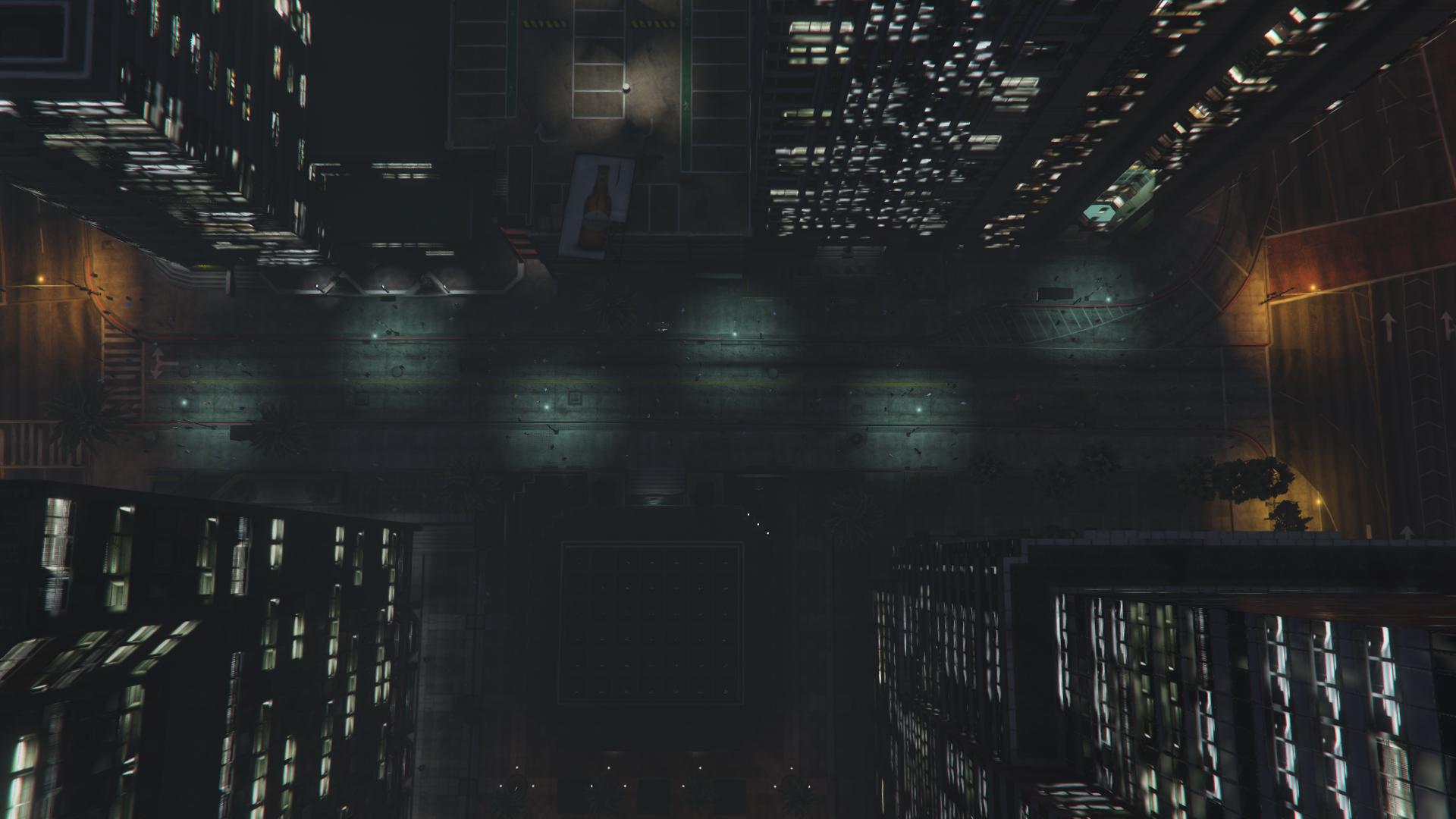}
        \end{minipage}

      \end{minipage}
  
      \caption{Some examples of the generated synthetic data.}
      \label{fig:sync}
  \end{figure}

  The cost of collecting and labeling real scene data is very high. Moreover, in general, the real scene is limited, and the diversity is weak. The standard data enhancement method can only alleviate this problem to a certain extent. To better increase the diversity of training data, we refer to \cite{wang2019learning}'s method and generate some synthetic data to assist training. Figure~\ref{fig:sync} demonstrates some of the synthetic data that we generated.
  
\section{Experiments}
  \label{sec:exp}
  \subsection{Dataset}
  VisDrone crowd count dataset (DroneCrowd) \cite{wen2019drone} is a remote crowd counting dataset with manual labels and 1080p RGB images collected by drones. 
  It consists of 112 sequences, of which 82 contains publicly reached point labels, and the rest of the sequences can only be tested by uploading the results to the official challenge \cite{zhu2020vision} server. 
  Therefore, we randomly divide the labeled 82 sequences into 75 for training and 7 for validation. 
  Our model achieves MAE 12.7 in the test server and wins the challenge\textsuperscript{\ref{final}}. 
  Because there is a limit on the number of test times, part of our following experimental results are based on the validation set.

  \subsection{Implementation Details}
  All experiments are build based on one NVIDIA TITAN RTX GPU and Intel(R) Xeon(R) Silver 4110 CPU @ 2.10GHz under Ubuntu 18.04 LTS operating system and PyTorch \cite{paszke2017automatic} framework.
  The ResNeSt50 encoder is initialized with the weights pre-trained on the ImageNet dataset, all other convolutional layers are initialized with normally distributed weights and zero bias. The learning rate is set to $10^{-5}$.

  \subsection{Performance Measures}

  \begin{table}[t]
    \center
    \caption{Results on the test set of several methods\textsuperscript{\ref{final}}.}
    \begin{threeparttable}
      \begin{tabular}{c|c|c|c|c}
        \hline
        \textbf{Method}                       &\textbf{MAE}                     &\textbf{User}  & \textbf{Institution}                           &  \textbf{Comment} \\ \hline
        MCNN \cite{zhang2016single} \tnote{*} \ \ & 34.7                            & -          & -                                     & -                                     \\
        MSCNN \cite{zeng2017multi} \tnote{*} \ \ & 58                              & -          & -                                     & -                                     \\
        CSRNet \cite{li2018csrnet} \tnote{*} \ \ & 19.8                            & -          & -                                     & -                                     \\
        `sa\_tta'  \tnote{**}        & 13.81         & l xl       &                                & Ranked Third                          \\
        `CSRNet' \tnote{**}         & 13.80          & shinan liu & beijing jiaotong univerisity   & Ranked Second                          \\  \hline
        Ours               & \textbf{12.70} & T Xini     & NWPU                           & \textbf{Ranked First}                         \\ \hline
        \end{tabular}
    \begin{tablenotes}
      \scriptsize
      \item[*] Results from official paper \cite{wen2019drone}.  
      \item[**] MAE from official test server \cite{visdroneWeb}. Data collected on 02:41:20, July 15, 2020, GMT. 
    \end{tablenotes}
    \end{threeparttable}
    \label{tab:res_test}
  \end{table}

  Researchers use MAE and mean squared error (MSE) to measure the differences between the predicted density map and the ground truth.
  The definition of MAE and MSE are given by following equations:
  \begin{equation}\begin{array}{l}
    \mathrm{MAE}=\frac{1}{H\times W} \sum_{i=1}^{H} \sum_{j=1}^{W}\left|z_{i, j}-\hat{z}_{i, j}\right|, \\
    \\
    \mathrm{MSE}=\sqrt{\frac{1}{H\times W} \sum_{i=1}^{H} \sum_{j=1}^{W}\left|z_{i, j}-\hat{z}_{i, j}\right|^{2}},
    \end{array}\end{equation}
  where $H$ and $W$ are the height and width of the test density map, $z_{i, j}$ is the ground truth pixel value of location $(i,j)$, $\hat{z}_{i, j}$ is the corresponding one of the predicted density map.
  
  The official challenge test server only provides MAE values. As for our experimental analysis, both MAE and MSE are calculated on the validation set.
  Table~\ref{tab:res_test} shows MAE values on the test set of several methods from the challenge \cite{zhu2020vision} and the official paper \cite{wen2019drone}.
  Our method outperforms all those methods and wins the challenge with MAE 12.70\textsuperscript{\ref{final}}.

  \subsection{Ablation Studies}
    \label{sec:abl}
    \ \ \ \ \ \textbf{Influence of Flow Stream: }
    To verify whether the flow stream helps improving the network, we set up experiments to train the network without the flow stream.
    We use different methods to generate the optical flow; the first one is based on deep learning \cite{sun2018pwc}, while the second one is based on dense inverse search \cite{kroeger2016dis}. We abbreviated them as PWC and DIS, respectively.
    We also take different combinations of $[f_x,f_y,f_h,f_s,f_{sub}]$ into consideration.
    Experiment results are shown in Table~\ref{tab:abl_flow}.
    Due to time limitations, we only train 30 epochs under each setting and take the best result.

    \begin{table}[t]
      \centering
      \caption{Results on the validation set of different types and combinations of the flow stream.}
      \begin{tabular}{c|c|c|c|c}
      \hline
      Stream         & Flow Type            & Flow Combination     & MAE (val)            & MSE (val)              \\ \hline
      Image          & -                    & -                    & 28.73                & 42.16                 \\
      \ Image+Flow \ & PWC                  & $[f_x,f_y,f_{sub}]$  & 26.36                & 36.10                 \\
      Image+Flow     & DIS                  & $[f_x,f_y,f_{sub}]$  & 26.31                & 38.26                 \\
      Image+Flow     & PWC                  & $[f_h,f_s,f_{sub}]$  & 23.14                & 31.09                 \\ \hline  \hline
      Image+Flow     & DIS                  & $[f_h,f_s,f_{sub}]$  & \textbf{22.66}       & \textbf{34.46}        \\
      \hline
      \end{tabular}
      \label{tab:abl_flow}
    \end{table}
    It can be seen that the introduction of the flow stream improves the generalization ability of the model, and its type and composition will also affect the final result.
    Optical flow generated by DIS, with combination method $[f_s,f_v,f_{sub}]$ achieves the best.

    \textbf{Influence of Gamma Correction: } 
    There are some extremely dark scenes in the test set, but there is no scene with such low light conditions in the training set. 
    These night scenarios have a significant impact on the test results. 
    The validation set can not verify the effectiveness of gamma correction.
    We submit results with/without gamma correction to the official test server.
    The test results are given in Table~\ref{tab:abl_gam}.
    \begin{table}[t]
      \centering
      \caption{Results on the validation set of different light condition enhancement sets.}
      \begin{tabular}{c|c|c}
        \hline
        Methods                  & \ \ MAE (val) \ \ & \ \ MSE(val) \ \ \\ \hline
        \ Without Gamma Correction\  & 28.37 & 39.42 \\ \hline
        With Gamma Correction    & \textbf{22.66}  &\textbf{34.46} \\ \hline
      \end{tabular}
      \label{tab:abl_gam}
    \end{table}
    It can be seen that random brightness transformation is performed on the input image during training to increase the diversity of the lighting conditions in the training data, which effectively improves the model's performance in the test set with low light environments.

    \textbf{Influence of Scale Variation: } 
    To simulate different cruising altitudes of the data collected by drones, we set experiments under the original image, scale input image between [0.7, 1.2], and [0.6, 1.8] randomly.
    Experiment results are shown in Table~\ref{tab:abl_sca}, the scale variation during the training process improved the model performance.
    According to the validation results, a broader range of scale change brings better results.
    However, if we apply scale change larger than [0.6, 1.8], the changed image size cannot meet the size requirements of the crop operation, so no more experiments are performed here.

    \begin{table}[t]
      \caption{MAE on the validation/test set under different training scale range and data.}
      \centering
      \begin{tabular}{c|c|c|c}
        \hline
        Scale Range    & \ \ MAE (val)\ \  & \ \ MSE (val)\ \   &\ \ MAE (test)\ \  \\ \hline
        None           & 36.55                & 47.65           & 24.61  \\
        {[}0.7, 1.2{]} & 29.19                & 41.58           & -   \\
        {[}0.6, 1.8{]} & \textbf{22.66}       & \textbf{34.46}  & 17.04   \\ \hline
        {[}0.6, 1.8{]} With Synthetic & -       & -  & \textbf{12.70}   \\ \hline
      \end{tabular}
      \label{tab:abl_sca}
    \end{table}

\begin{figure}[tbp]
      \centering
      \begin{minipage}[t]{0.98\textwidth}
  
        \begin{minipage}[t]{0.328\textwidth}
          \centering
          \includegraphics[width=1\textwidth]{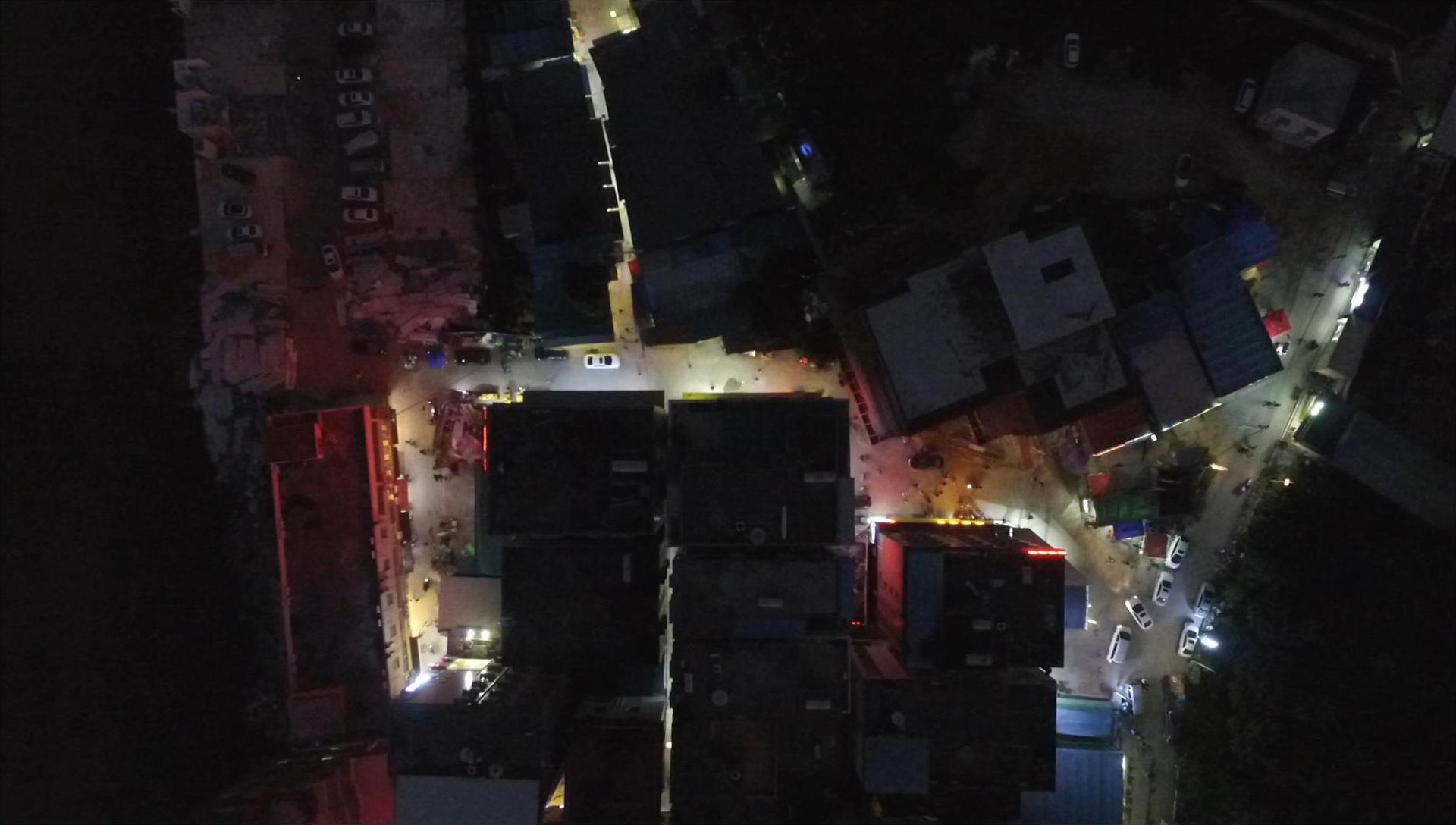}
          Original Image
        \end{minipage}
        \begin{minipage}[t]{0.328\textwidth}
          \centering
          \includegraphics[width=1\textwidth]{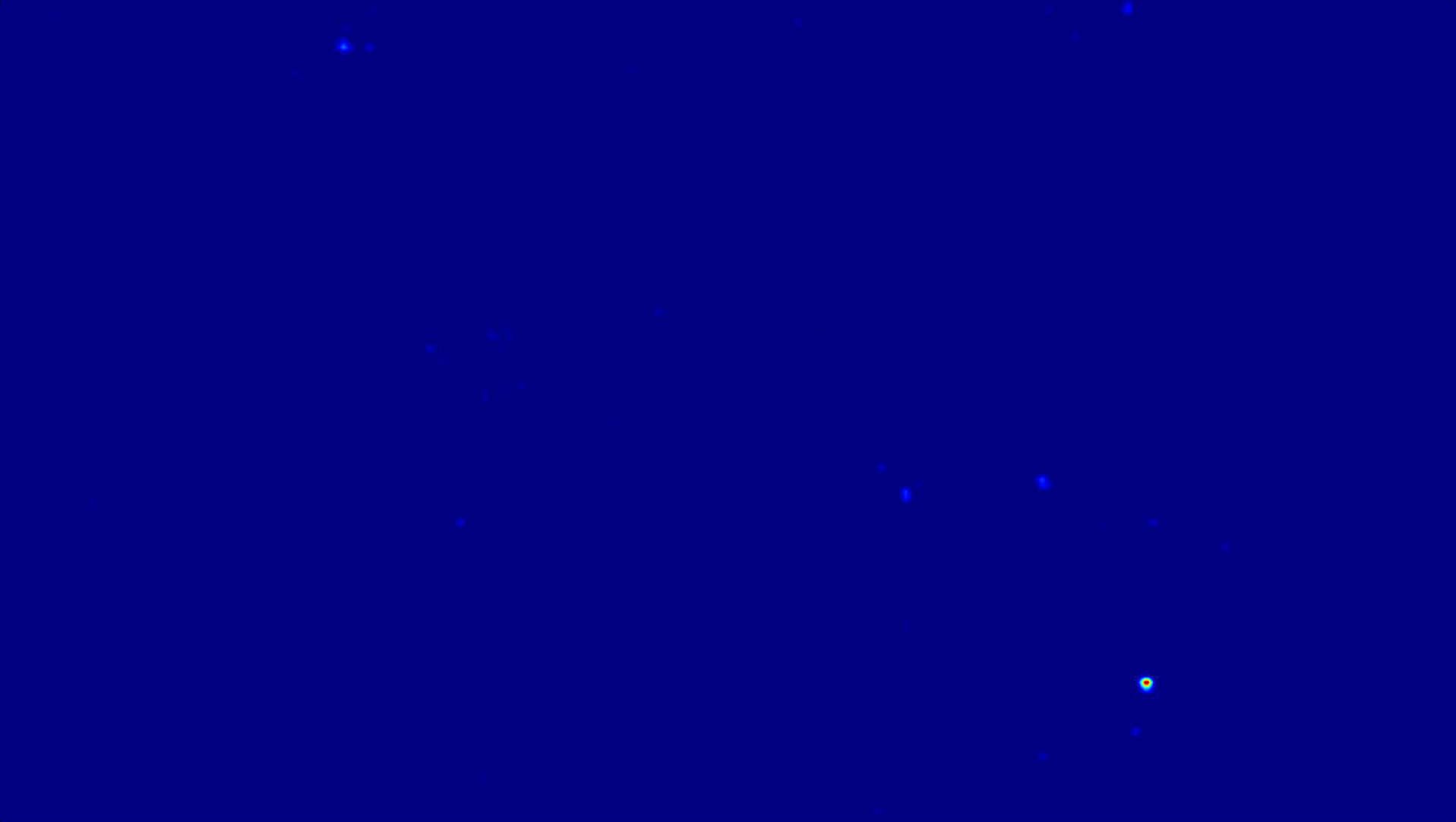}
          Result of model train under normal data
        \end{minipage}
        \begin{minipage}[t]{0.328\textwidth}
          \centering
          \includegraphics[width=1\textwidth]{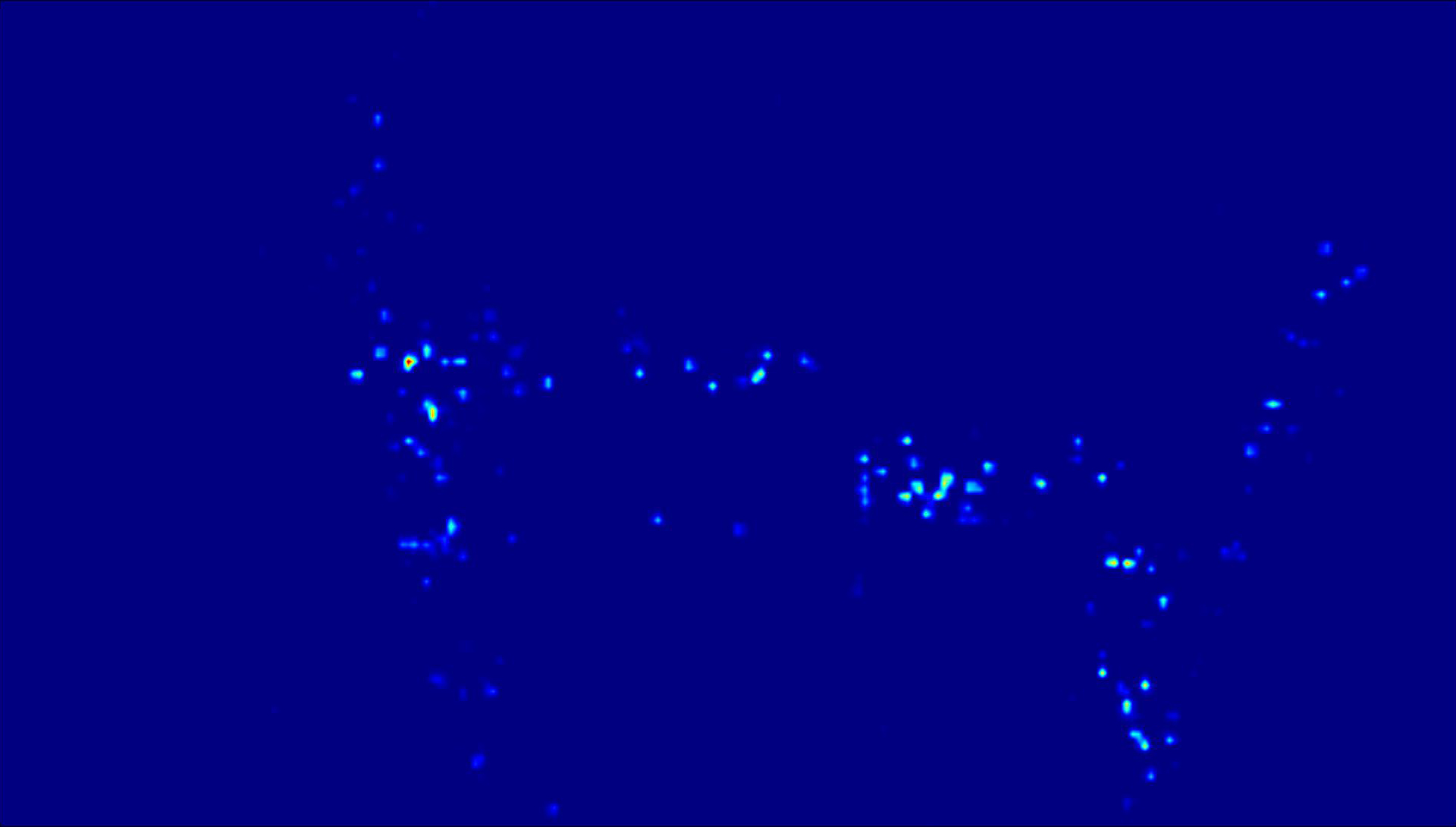}
          Result of model train under synthetic data
        \end{minipage}

      \end{minipage}
  
      \caption{Visualization of density map generate by different models.}
      \label{fig:syncres}
  \end{figure}

    \textbf{Influence of Synthetic Data: } 
    We use the synthetic data alone to train a new model to predict the night scene, and then use the original model to predict the daytime scene. Table~\ref{tab:abl_sca} also shows the results of mixing the two models. It can be seen that the synthetic data has brought a significant improvement to the final result.
    Figure~\ref{fig:syncres} shows the visualization of the predict density map under normal training data or synthetic training data.
    It is clear that the model trained under ordinary data can not make good predictions in extremely dark scenes.
    Nevertheless, the model trained under synthetic data can give good results.

\section{Conclusion}
  This paper proposes a bi-path optical flow-based crowd counter network, together with several data argumentation methods specially designed for remote crowd count tasks.
  To be specific, two separate ResNeSt50 based encoder-decoder stream extract feature vectors from RGB images and 3-channel optical flow tensors.
  During the training phase, data argumentation operations like crop, random flip, gamma correction, scale variation are applied to create nonexistent scenes from the train set.
  Synthetic data are generated for better performance under extreme dark scenarios.
  With the combination of network design, data expansion and synthetic data, we finally win the official competition\textsuperscript{\ref{final}}.
  In addition, we designed sufficient experiments to verify and analyze the effectiveness of network structure design and different data argumentation methods.
  In the future, we will explore effective network design and data argumentation methods for remote crowd count task.

\clearpage
%
%
\bibliographystyle{splncs04}
\bibliography{egbib}
\end{document}